\documentclass{article}

\usepackage[numbers, compress]{natbib}

\usepackage{xcolor}
\usepackage{soul}

\usepackage[final]{neurips_2019}

\usepackage[utf8]{inputenc} 
\usepackage[T1]{fontenc}    
\usepackage{hyperref}       
\usepackage{url}            
\usepackage{booktabs}       
\usepackage{amsfonts}       
\usepackage{nicefrac}       
\usepackage{microtype}      
\usepackage{graphicx}
\usepackage{multirow}
\title{Fraud detection in telephone conversations for financial services using linguistic features}

\author{%
  Nikesh~Bajaj, Tracy~Goodluck~Constance, Marvin~Rajwadi, Julie~Wall, Mansour~Moniri
  \\
  Intelligent Systems Research Group, University of East London, UK\\
  \texttt{\{n.bajaj, t.goodluckconstance, m.rajwadi, j.wall, m.moniri\}@uel.ac.uk} \\
   \AND
   Cornelius~Glackin, Nigel~Cannings\\
   Intelligent Voice Ltd., London, UK\\
   \texttt{\{neil.glackin, nigel.cannings\}@intelligentvoice.com}
   \And
   Chris~Woodruff, James~Laird\\
   Strenuus Ltd., London, UK\\
  \texttt{\{chris.woodruff, james.laird\}@strenuusltd.com}
}

\begin{document}    
\maketitle
\begin{abstract}
 Detecting the elements of deception in a conversation is one of the most challenging problems for the AI community. It becomes even more difficult to design a transparent system, which is fully explainable and satisfies the need for financial and legal services to be deployed. This paper presents an approach for fraud detection in transcribed telephone conversations using linguistic features. The proposed approach exploits the syntactic and semantic information of the transcription to extract both the linguistic markers and the sentiment of the customer's response. We demonstrate the results on real-world financial services data using simple, robust and explainable classifiers such as Naive Bayes, Decision Tree, Nearest Neighbours, and Support Vector Machines.
\end{abstract}
\section{Introduction}
With the rapid increases in technological development, fraud is a major concern for safe and trustworthy digital communications and transactions. To deal with this, several data mining and machine learning systems have been developed \cite{ngai2011application}. The techniques available in the literature have mostly focused on financial fraud, such as insurance and credit card fraud \cite{buanuarescu2015detecting}, analysing the anomalous behaviour of the customer in financial transactions \cite{zareapoor2015application}. However, fraud is not only limited to financial transactions, it also involves being deceptive by providing false statements, such as in loan applications and when trying to uncover the account status of others, etc. 

Compared to other fraud detection techniques, we have not yet found other research in the literature which reports the analysis of telephone conversations between customers and service providers, such as insurance companies or banks. In fact, a telephone conversation is usually the first point of contact. Analysis of this data-rich communication can reveal potential fraudulent cues at a very early stage to prevent fraud. To achieve this, a linguistic based approach using Natural Language Processing (NLP) techniques \cite{manning1999foundations} can be used. 

The requirement for financial and legal services to deploy any advanced Artificial Intelligence (AI) algorithm is that it must be a transparent and fully explainable system. Financial institutions are required to explain and justify the decision taken for customers or clients, for example, whether they will pay out on an insurance claim or not. This limits the use of sophisticated and complex systems such as 
Deep Neural Networks (DNN) in this area. However, the most recent advancements are tending towards achieving full Explainable AI \cite{gunning2017explainable, ribeiro2016should, lundberg2017unified}.

In this paper, we propose an approach to use linguistic features by exploiting the syntactic and semantic information in the transcripts of telephone conversations. We demonstrate the results of this approach on real-world data, collected from two financial services institutions. We trained simple, robust and explainable classifiers to achieve an explanation of the decision process while revealing the importance of features responsible for the decision.

\section{Proposed Approach}
The proposed approach is designed to analyse the transcription of a telephone conversation, generated using state-of-the-art Automatic Speech Recognition technology in real-time, which allows our deception detection approach to also work in real-time. This approach extracts the linguistic features of the transcription and trains the explainable classifiers to analyse and validate the decision process. 
\label{ss:lingC}

Language is a medium of communication, where the choice of words can reflect the emotional and cognitive state of the speaker. Only training and rehearsal can allow a speaker to control their vocabulary to not leak any emotional state \cite{henry1936linguistic}. Psychologists suggest that speakers often have no control over their choice of words and can reveal their emotions involuntarily \cite{schafer2007grammatical}. Deceptive speech is considered to be a deliberate choice to mislead and the language used can reveal an underlying cognitive state. Many studies have shown that linguistic cues can indicate the elements of deception in language \cite{pennebaker2001linguistic, cohen2014detecting}. This work considers two types of linguistic features. One is extracted from the syntax of the language, focusing primarily on the words used, called \textit{Linguistic Markers}. The second focuses on the overall sentence structure, its semantics, and the sentiment of the dialogue.

\subsection{Linguistic Markers}
\label{sss:markers}
The observation of linguistic cues to detect deceptive speech focuses mainly on word/phrases either in oral or written form. There is an exhaustive list of linguistic markers provided by \cite{pennebaker2001linguistic, humpherys2010system}, and we have adapted a subset of these linguistic markers for our study, provided in Table \ref{T:lMarker}.
\begin{table}
\small
\centering
\caption{Linguistic Markers}
\label{T:lMarker}
\begin{tabular}{p{8.5cm} p{4.5cm}}
\hline
\toprule
Marker & Example  \\ [0.5ex] 
\hline
\toprule
\textbf{Causation:}  Providing a certain level of concreteness to an explanation. \cite{pennebaker2001linguistic,hancock2005automated} & Because, Effect, Hence	 \\
\textbf{Negation:} Avoiding to provide a direct response \cite{adams2002communication} & No, Not, Can't, Didn't \\
\textbf{Hedging:} Describes words which meaning implicitly involves fuzziness \cite{bachenko2008verification} & May be, I guess, Sort of\\
\textbf{Qualified assertions:} Unveils questionable actions \cite{bachenko2008verification} & Needed, Attempted\\
\textbf{Temporal Lacunae:} Unexplained lapses of time \cite{bachenko2008verification} & Later that day, Afterwards\\
\textbf{Overzealous expression:} Expresses some level of uncertainty \cite{bachenko2008verification} & I swear to God, Honestly\\
\textbf{Memory loss:} Feigning memory loss \cite{bachenko2008verification} & I forget, Can’t remember\\
\textbf{Third person plural pronouns:} Possessive determiners to refer to things or people other than the speaker \cite{pennebaker2001linguistic} & They, Them, Theirs\\
\textbf{Pronouns:} Possessive determiners to refer to the speaker by overemphasising their physical presence \cite{pennebaker2001linguistic,li2014towards}& I, Me, Mine\\
\textbf{Negative emotion:} Negative expressions in word choice \cite{skillicorn2010patterns,pennebaker2001linguistic,jurafsky2000speech1} & Afraid, Sad, Hate, Abandon, Hurt\\
\textbf{Negative sentiment:} Negative emotional effect \cite{jurafsky2000speech1} & Abominable, Anger, Anxious, Bad\\
\textbf{Positive emotion:} Positive expressions in word choice \cite{pennebaker2001linguistic,jurafsky2000speech1} & Happy, Brave, Love, Nice, Sweet\\
\textbf{Positive sentiment:} Positive emotional effect \cite{jurafsky2000speech1} & Admire, Amazing, Assure, Charm\\
\textbf{Disfluencies:} Interruption in the flow of speech \cite{pennebaker2001linguistic} & Uh, Um, You know, Er, Ah\\
\textbf{Self reference words:} Deceivers tend to use fewer self-referencing expressions  \cite{bachenko2008verification} & I, My, Mine\\
\textbf{Nominalised verbs:} Nouns derived from verbs. Nominalisations tend to hide the real action.  \cite{lapata2002disambiguation} & Education, Arrangement
\\
\hline
\toprule
\end{tabular}
\end{table}

\subsection{Sentiment}
Many studies support the presence of negative words and sentiment in deceptive speech \cite{rudacille1994identifying, watson1981oral, wade1993communicating}, which can be spotted by linguistic markers. However, linguistic markers analyse only the syntactic information of the dialogue text and are prone to miss the overall sentiment. In cases where the sentence has a negative sentiment but with no existing negative words, linguistic markers have limited capability. To overcome this, we use sentiment as a feature, which reflects the polarity of the speaker's feelings in a dimension from negative to positive. Therefore, sentiment can be used to detect the elements of deception in speech. To extract the sentiment of dialogue in telephone conversations, we used a DNN which was trained on the IMDB movie review dataset \cite{maas2011learning}. Full information on the development, training, and evaluation of this sentiment-analysis based DNN can be found in our previous work \cite{marvin@2019_11}. The use of a trained model on IMDB's dataset is considered to be effective, assuming that the set of domain-dependent words is very small \cite{yoshida2011transfer}. However, an efficient domain adaptation using transfer learning can be used to extract the sentiment for given context \cite{calais2011bias}.

\section{Experiments \& Results}
\label{S:exp}
\subsection{Dataset}
\label{ss:data}
To evaluate our proposed approach, we employed real-world data collected from financial services institutions. This dataset contains the transcriptions of 56 telephone conversations, collected from two different financial institutions. Ideally, a larger dataset would be desirable, however, this data is limited volume due to legal \& ethical constraints. The transcription of each conversation includes the operator's questions and the customer's responses. From the dataset of 56 calls, 32 are known fraudulent calls and 24 are non-fraudulent. As this is real-world data, the timing of the calls varies widely. The average number of responses are $19 \pm 15$. The shortest conversation in the dataset has only four responses from the customer; while the longest has 101 responses.

\subsection{Feature Extraction and Modeling}
From each telephone conversation, only the customer's responses are used for the feature extraction. Two types of linguistic features are extracted: namely \textit{Linguistic Markers} and \textit{Sentiment}, as explained in Section \ref{ss:lingC}. For linguistic markers, the frequency of each of the 16 markers from Table \ref{T:lMarker} present in the customer's response is computed. Then the sentiment of each customer response is estimated using the DNN and scaled from -1 to 1. As there are a different number of responses in each conversation, we computed statistical measures from each individual response within the conversations. In total, 11 sentiment-related features were extracted, namely: mean, standard deviation (SD), minimum (Min), maximum (Max), median, interquartile range (IQR), Kurtosis, Skewness, positive energy (pE), negative energy (nE), and total number of responses (tR).  

\begin{figure}[h!]
  \centering
  \includegraphics[width=1\textwidth]{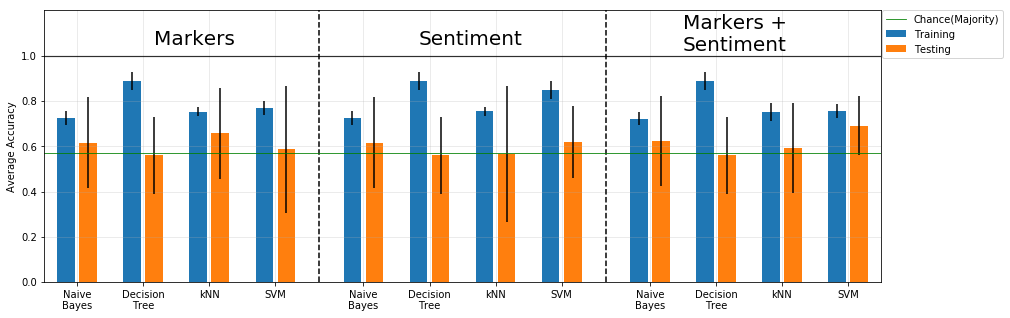}
  \caption{Average performance of K-Fold(K=10) for different models}
  \label{fig:res}
\end{figure}

Finally, we trained the models (Naive Bayes, Decision Tree (DTree), k-Nearest Neighbors (kNN), and Support Vector Machines (SVM) with the individual features (e.g. marker, sentiment) and then we combined the features and re-trained. The parameters for the models are as follows: DTree with $depth=3$, kNN with $k=3$, and SVM with linear kernel and $C=1$. The choice of models and their parameters were restricted by three properties: simplicity, robustness, and explainability. Since the data size is small, we trained and tested each model with K-Fold cross-validation, with K=10. The mean and SD of the training and testing accuracies for each model are tabulated in Table \ref{T:results} and plotted in Figure \ref{fig:res}.



\subsection{Discussion}
\begin{figure}[h!]
  \centering
  \includegraphics[width=0.95\textwidth]{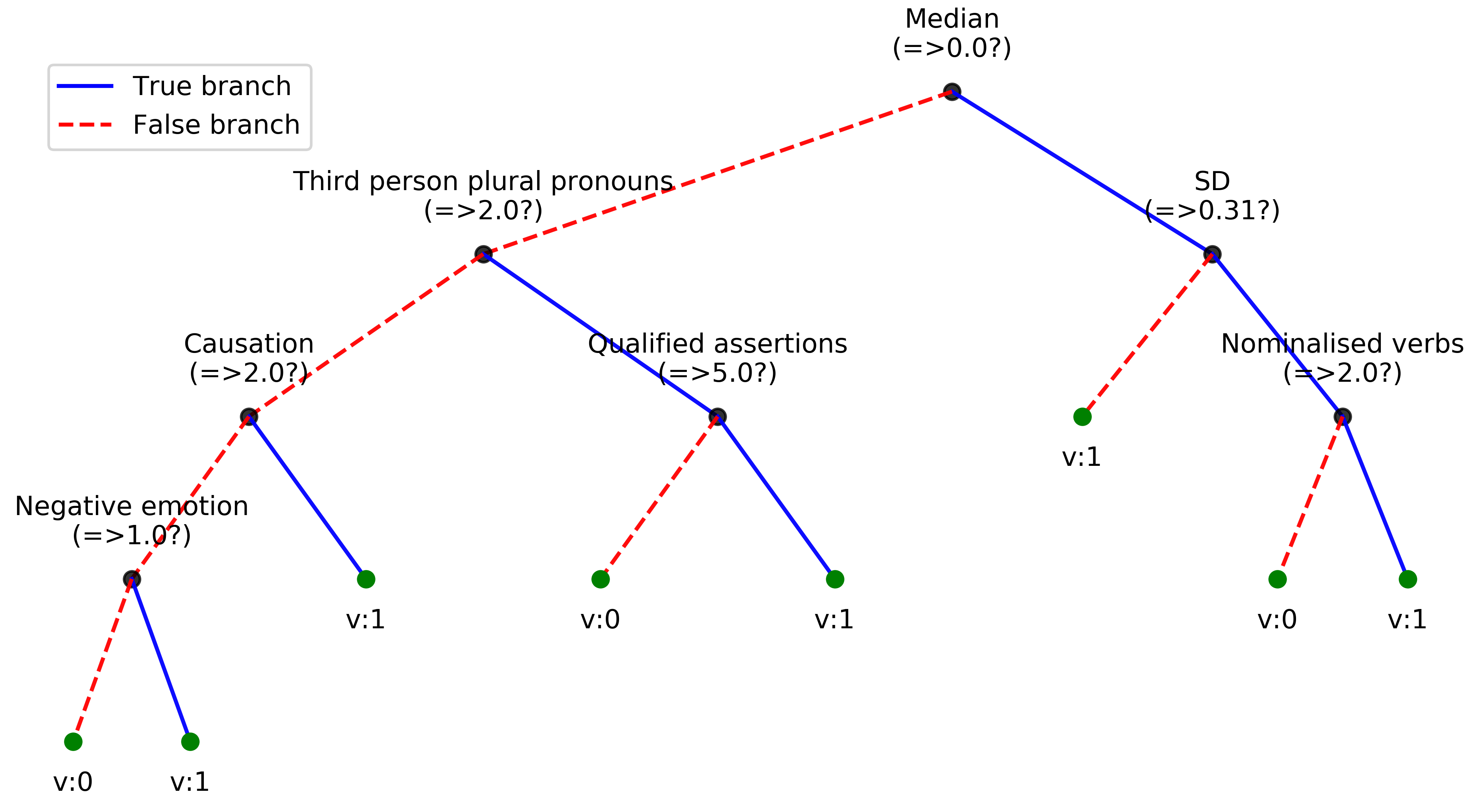}
  \caption{A Decision Tree for fraud detection. Leaf node v:0 - Non-Fraud, v:1 - Fraud}
  \label{fig:DTree}
\end{figure}

It can be observed that the highest testing accuracy achieved with solely the linguistic markers is 65.5\% using kNN, whereas, with sentiment features, it is only 62\% using an SVM. However combining both features, improves the testing accuracy to 69\% with only 0.1 deviation for the SVM. One of the decision trees built with combined features is shown in Figure \ref{fig:DTree}. The decision process is very visible from the tree and shows the importance of the features. For example, the most important feature is - Median of sentiment value with threshold 0. Another important feature is 'third person plural pronoun', which is indication of deception, reflecting a customer's attempt to discuss third person, while the call is about his/her own financial account. It can also be noticed that qualified assertion, negative emotion, causation, and nominalised verb are also important linguistic features. Interestingly, a variation in sentiment values (SD) of responses, is also an important feature, indicating the too much change in the language of customer.

\begin{table}[h!]
\small
 \centering
 \caption{Results of modeling with K-Fold (K=10)}
 \label{T:results}
 \begin{tabular}{l l l l l l}
 \hline
 \toprule
 \multirow{2}{4em}{Features} & \multirow{2}{4em}{Accuracy}  & \multicolumn{4}{c}{Model} \\[0.5ex] 
  &   &Naive Bayes & DTree (d=3)& kNN(k=3) & SVM(Linear) \\ [0.5ex] 
 \hline
 \toprule
 \multirow{2}{4em}{Markers}
 & Training & $0.7241 \pm0.03$ & $0.8871 \pm0.04$ & $0.7521 \pm0.02$  & $0.7679 \pm0.03$\\
 & Testing  & $0.6167 \pm0.20$ & $0.5600 \pm0.17$ & $0.6567 \pm0.20$  & $0.5867 \pm0.28$\\[0.3ex] 
 \midrule
 \multirow{2}{4em}{Sentiment} 
 & Training & $0.7241 \pm0.03$ & $0.8871 \pm0.04$ & $0.7540 \pm0.02$ & $0.8491 \pm0.04$\\
 & Testing  & $0.6167 \pm0.20$ & $0.5600 \pm0.17$ & $0.5667 \pm0.30$ & $0.6200 \pm0.16$\\[0.3ex]
 \midrule
 \multirow{2}{4em}{Markers + Sentiment} 
 & Training & $0.7222 \pm0.03$ & $0.8871 \pm0.04$ & $0.7500 \pm0.04$ & $0.7560 \pm0.03$\\
 & Testing  & $0.6233 \pm0.20$ & $0.5600 \pm0.17$ & $0.5933 \pm0.20$ & $0.6900 \pm0.13$\\[0.3ex]
\hline
\toprule
\end{tabular}
\end{table}

\section{Conclusions and future work}
\label{S:conclusion}
The proposed approach for fraud detection in financial services telephone conversations has employed two different types of linguistic features, namely markers and sentiment. While markers exploit the syntactic information of the conversation, sentiment uses semantic information. The results presented in the paper show that combining these features produces the highest average accuracy. In order to achieve transparency in the decision process, with the limited dataset size, the choice of models were kept simple, robust and explainable. The financial and legal services are required to explain the decision process made with any mode of operation. For this same reason, an example of a decision tree from these experiments is shown in Figure \ref{fig:DTree}. With a small decision tree, it is easy to explain the procedure producing outcome. Future work plans to extend the presented work for different scenarios including legal and insurance services, with the aim to again employ real-world data. 
\small
\bibliographystyle{unsrt}
\bibliography{reference}

\begin{thebibliography}{10}

\bibitem{ngai2011application}
Eric~WT Ngai, Yong Hu, Yiu~Hing Wong, Yijun Chen, and Xin Sun.
\newblock The application of data mining techniques in financial fraud
  detection: A classification framework and an academic review of literature.
\newblock {\em Decision support systems}, 50(3):559--569, 2011.

\bibitem{buanuarescu2015detecting}
Adrian B{\u{a}}n{\u{a}}rescu.
\newblock Detecting and preventing fraud with data analytics.
\newblock {\em Procedia economics and finance}, 32:1827--1836, 2015.

\bibitem{zareapoor2015application}
Masoumeh Zareapoor, Pourya Shamsolmoali, et~al.
\newblock Application of credit card fraud detection: Based on bagging ensemble
  classifier.
\newblock {\em Procedia computer science}, 48(2015):679--685, 2015.

\bibitem{manning1999foundations}
Christopher~D Manning, Christopher~D Manning, and Hinrich Sch{\"u}tze.
\newblock {\em Foundations of statistical natural language processing}.
\newblock MIT press, 1999.

\bibitem{gunning2017explainable}
David Gunning.
\newblock Explainable artificial intelligence (xai).
\newblock {\em Defense Advanced Research Projects Agency (DARPA), nd Web}, 2,
  2017.

\bibitem{ribeiro2016should}
Marco~Tulio Ribeiro, Sameer Singh, and Carlos Guestrin.
\newblock Why should i trust you?: Explaining the predictions of any
  classifier.
\newblock In {\em Proceedings of the 22nd ACM SIGKDD international conference
  on knowledge discovery and data mining}, pages 1135--1144. ACM, 2016.

\bibitem{lundberg2017unified}
Scott~M Lundberg and Su-In Lee.
\newblock A unified approach to interpreting model predictions.
\newblock In {\em Advances in Neural Information Processing Systems}, pages
  4765--4774, 2017.

\bibitem{henry1936linguistic}
Jules Henry.
\newblock The linguistic expression of emotion.
\newblock {\em American Anthropologist}, 38(2):250--256, 1936.

\bibitem{schafer2007grammatical}
John~R Schafer.
\newblock {\em Grammatical differences between truthful and deceptive written
  narratives}.
\newblock Citeseer, 2007.

\bibitem{pennebaker2001linguistic}
James~W Pennebaker, Martha~E Francis, and Roger~J Booth.
\newblock Linguistic inquiry and word count: Liwc 2001.
\newblock {\em Mahway: Lawrence Erlbaum Associates}, 71(2001):2001, 2001.

\bibitem{cohen2014detecting}
Katie Cohen, Fredrik Johansson, Lisa Kaati, and Jonas~Clausen Mork.
\newblock Detecting linguistic markers for radical violence in social media.
\newblock {\em Terrorism and Political Violence}, 26(1):246--256, 2014.

\bibitem{humpherys2010system}
Sean~L Humpherys.
\newblock A system of deception and fraud detection using reliable linguistic
  cues including hedging, disfluencies, and repeated phrases.
\newblock 2010.

\bibitem{hancock2005automated}
Jeffrey~T Hancock, Lauren Curry, Saurabh Goorha, and Michael Woodworth.
\newblock Automated linguistic analysis of deceptive and truthful synchronous
  computer-mediated communication.
\newblock In {\em Proceedings of the 38th Annual Hawaii International
  Conference on System Sciences}, pages 22c--22c. IEEE, 2005.

\bibitem{adams2002communication}
Susan~H Adams.
\newblock {\em Communication under stress: indicators of veracity and deception
  in written narratives}.
\newblock PhD thesis, Virginia Tech, 2002.

\bibitem{bachenko2008verification}
Joan Bachenko, Eileen Fitzpatrick, and Michael Schonwetter.
\newblock Verification and implementation of language-based deception
  indicators in civil and criminal narratives.
\newblock In {\em Proceedings of the 22nd International Conference on
  Computational Linguistics-Volume 1}, pages 41--48. Association for
  Computational Linguistics, 2008.

\bibitem{li2014towards}
Jiwei Li, Myle Ott, Claire Cardie, and Eduard Hovy.
\newblock Towards a general rule for identifying deceptive opinion spam.
\newblock In {\em Proceedings of the 52nd Annual Meeting of the Association for
  Computational Linguistics (Volume 1: Long Papers)}, pages 1566--1576, 2014.

\bibitem{skillicorn2010patterns}
David~B Skillicorn and Ayron Little.
\newblock Patterns of word use for deception in testimony.
\newblock In {\em Security Informatics}, pages 25--39. Springer, 2010.

\bibitem{jurafsky2000speech1}
Dan Jurafsky.
\newblock {\em Lexicons for Sentiment, Affect, and Connotation}.
\newblock Pearson Education India, 2000.

\bibitem{lapata2002disambiguation}
Maria Lapata.
\newblock The disambiguation of nominalizations.
\newblock {\em Computational Linguistics}, 28(3):357--388, 2002.

\bibitem{rudacille1994identifying}
Wendell~C Rudacille.
\newblock {\em Identifying lies in disguise}.
\newblock Kendall/Hunt, 1994.

\bibitem{watson1981oral}
Kittie~Wells Watson.
\newblock Oral and written linguistic indices of deception during employment
  interviews.
\newblock 1981.

\bibitem{wade1993communicating}
Elizabeth Wade.
\newblock {\em Communicating about narrative (in) accuracy}.
\newblock Stanford University, 1993.

\bibitem{maas2011learning}
Andrew~L Maas, Raymond~E Daly, Peter~T Pham, Dan Huang, Andrew~Y Ng, and
  Christopher Potts.
\newblock Learning word vectors for sentiment analysis.
\newblock In {\em Proceedings of the 49th annual meeting of the association for
  computational linguistics: Human language technologies-volume 1}, pages
  142--150. Association for Computational Linguistics, 2011.

\bibitem{marvin@2019_11}
Marvin Rajwadi, Cornelius Glackin, Julie Wall, G{\'e}rard Chollet, and Nigel
  Cannings.
\newblock Explaining sentiment classification.
\newblock {\em Proc. Interspeech 2019}, pages 56--60, 2019.

\bibitem{yoshida2011transfer}
Yasuhisa Yoshida, Tsutomu Hirao, Tomoharu Iwata, Masaaki Nagata, and Yuji
  Matsumoto.
\newblock Transfer learning for multiple-domain sentiment
  analysis—identifying domain dependent/independent word polarity.
\newblock In {\em Twenty-Fifth AAAI Conference on Artificial Intelligence},
  2011.

\bibitem{calais2011bias}
Pedro~Henrique Calais~Guerra, Adriano Veloso, Wagner Meira~Jr, and
  Virg{\'\i}lio Almeida.
\newblock From bias to opinion: a transfer-learning approach to real-time
  sentiment analysis.
\newblock In {\em Proceedings of the 17th ACM SIGKDD international conference
  on Knowledge discovery and data mining}, pages 150--158. ACM, 2011.

\end{thebibliography}

\end{document}